\PassOptionsToPackage{numbers,square,sort&compress}{natbib}
\documentclass{article}
\usepackage{arxiv}
\usepackage{amsmath,amsfonts}
\usepackage{array}
\usepackage{booktabs}
\usepackage{graphicx}
\usepackage{makecell}
\usepackage{multirow}
\usepackage[table]{xcolor}
\usepackage{hyperref}
\hypersetup{
	colorlinks=true,
	linkcolor=blue,
	citecolor=blue,
	urlcolor=blue
}

\usepackage{url}
\usepackage{lineno}
\def\BibTeX{{\rm B\kern-.05em{\sc i\kern-.025em b}\kern-.08em
		T\kern-.1667em\lower.7ex\hbox{E}\kern-.125emX}}

\providecommand{\IEEEPARstart}[2]{#1#2}
\renewcommand{\markboth}[2]{}

\begin{document}
	
	\title{Meta-Learning-Assisted Constraint Relaxation for Constrained Black-Box Optimization}
	\author{Sijie Ma \quad Zeyuan Ma \quad Yue-Jiao Gong \quad Ran Cheng}

	\author{
        \textbf{Sijie~Ma}$^{1}$,
        \textbf{Zeyuan~Ma}$^{2,}$\thanks{Zeyuan Ma is the corresponding author (mzy@ieee.org).}~,
        \textbf{Yue-Jiao~Gong}$^{3}$,
        \textbf{Ran~Cheng}$^{1}$\\
        [-1pt]
        $^{1}$Hong Kong Polytechnic University \quad $^{2}$South China Normal University\\
        $^{3}$South China University of Technology
	}
    
	\maketitle
	
	\begin{abstract}
		Constraint handling is central to constrained black-box optimization (BBO), where objective improvement and feasibility restoration often provide conflicting search signals. Existing $\epsilon$-relaxation methods are simple and effective, but their relaxation schedules are usually fixed or manually designed for a limited range of problems. To address this limitation, this letter proposes MeCO, a meta-learning-assisted optimizer that learns an adaptive $\epsilon$-relaxation policy for constrained BBO. MeCO couples a SHADE optimizer with a Double Deep Q-Network controller. At each optimization step, the controller observes compact population and constraint features and selects a scalar action, which is decoded into a relaxation vector for the candidate comparison rule. The policy is trained across constrained BBO instances and then deployed on held-out problems without problem-specific tuning. Experiments on the CEC2017 constrained benchmark, 16 UAV path-planning tasks and eight real-world engineering problems provide evidence that MeCO transfers across held-out benchmark functions, higher dimensions, and an application-domain setting. Ablation and behavior analyses further clarify the roles of constraint-related state features, action scaling, reward shaping, and meta-training.
		
	\end{abstract}
    
    \keywords{Black-Box Optimization \and Constraint Handling \and Dynamic Algorithm Configuration \and Reinforcement Learning}
    
	\section{Introduction}
	\IEEEPARstart{C}{onstrained} black-box optimization (BBO) appears in engineering design and process optimization, where the objective and constraint functions can only be queried through evaluations~\cite{2jones2008large2,3boukouvala2017global3}. The optimizer must find feasible solutions and improve their objective values within a limited evaluation budget. If feasibility pressure is too strong, the search may settle around mediocre feasible points. If feasibility pressure is too weak, many evaluations may be spent in infeasible regions. Consequently, constraint handling impacts the behavior of population-based constrained BBO~\cite{coello-1,coello-2} such as Differential Evolution~(DE)~\cite{storn1997differential} algorithms and Evolution Strategy~(ES)~\cite{schwefel1977evolutionsstrategien} algorithms.
	
	\begin{figure}[!t]
		\centering
		\includegraphics[width=0.65\linewidth]{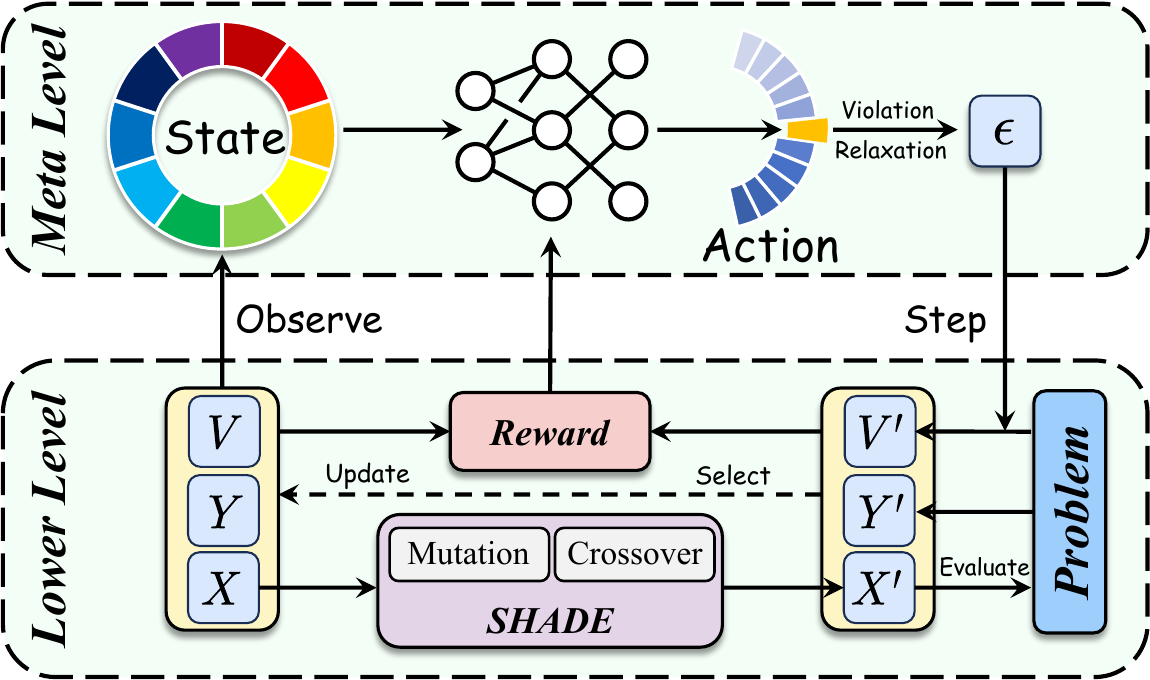}
		\caption{Bi-level workflow of MeCO. The SHADE optimizer searches with an $\epsilon$-relaxation rule, while the policy sets the relaxation action from population features.}
		\label{fig:workflow}
	\end{figure}
	
	Existing constraint-handling methods often require problem-dependent design choices. Penalty methods, multi-objective transformations, and surrogate-assisted schemes are common choices~\cite{coello-2,regis2019survey}. They can be effective, but they also introduce penalty coefficients, selection mechanisms, or surrogate-modeling decisions. $\epsilon$-relaxation is lighter because it treats mildly infeasible candidates as acceptable during comparison, thereby allowing the search to use nearly feasible regions as stepping stones~\cite{takahama2010constrained,takahama2006constrained, 9185566}. However, the relaxation threshold still needs a schedule: fixed values require problem-dependent choices, while handcrafted adaptive rules can encode problem-specific bias~\cite{fan2017improved,yang2014epsilon}.
	
	Recent MetaBBO studies show that optimizer behavior can be improved by learning control policies over a distribution of problems~\cite{metabbosurvey,rl-bbo-survey}. Existing RL-based MetaBBO methods mainly configure generic optimizer components, such as operator selection~\cite{rldas}, parameter control~\cite{rldeafl,configx}, and update-rule generation~\cite{10metabox10}. However, constraint-handling control remains less explored. This gap is important because the useful level of feasibility pressure changes during a run: early search may benefit from accepting nearly feasible candidates, while later search often needs stricter feasibility restoration. Thus, a learned controller can use online search signals to adjust feasibility pressure, instead of following a relaxation schedule chosen before the run.
	
	This letter studies whether the relaxation policy itself can be learned from a distribution of constrained BBO problems. To this end, we propose a \textbf{\underline{Me}}ta-learning-assisted optimizer for \textbf{\underline{C}}onstrained Black-Box \textbf{\underline{O}}ptimization, termed \textbf{MeCO}. MeCO treats $\epsilon$ control as dynamic algorithm configuration. A SHADE optimizer~\cite{tanabe2013success} performs the search, while a reinforcement learning policy~\cite{rl} observes the population state and selects a relaxation action for candidate selection. The policy does not replace the search operator or model the objective function; instead, it controls the feasibility pressure used in the comparison rule. Once trained, the policy is reused on held-out problems without problem-specific tuning.
	
	The contributions are threefold. First, MeCO formulates adaptive $\epsilon$-relaxation control as a meta-level Markov decision process (MDP), which separates lower-level search from upper-level feasibility-pressure control. Second, MeCO defines compact state, action, and reward designs that describe feasibility progress and objective-constraint tradeoffs without depending on the number of constraints or the problem dimension. Third, experiments on CEC2017, UAV path planning and real-world engineering problems provide evidence that the learned policy transfers competitively against representative constrained BBO baselines under the tested protocols.

	\section{Methodology}
	\label{sec:method}
	\subsection{Constrained BBO and Relaxation}
	The lower-level optimizer solves a constrained minimization problem:
	\begin{equation}
		\begin{aligned}
			\min_{\mathbf{x}} \quad &f(\mathbf{x}), \quad \mathbf{x}^L \leq \mathbf{x} \leq \mathbf{x}^U,\\
			\mathrm{s.t.} \quad &g_i(\mathbf{x}) \leq 0,\ i=1,\ldots,p, \\ 
			& h_j(\mathbf{x})=0,\ j=1,\ldots,q.
		\end{aligned}
		\label{eq:cop}
	\end{equation}
	The standard aggregated violation is
	$\nu(\mathbf{x})=\sum_i\max\{g_i(\mathbf{x}),0\}+\sum_j |h_j(\mathbf{x})|$.
	This definition is also used for final performance evaluation, so relaxation affects only the search process rather than the reported metric.
	In $\epsilon$-relaxation, the comparison rule uses a relaxation vector $\boldsymbol{\epsilon}_t=[\epsilon_{t,1},\ldots,\epsilon_{t,p+q}]^\top$ and computes the relaxed violation as
	\begin{equation}
		\begin{aligned}
			\nu_{\boldsymbol{\epsilon}}(\mathbf{x})=&
			\sum_i\max\{g_i(\mathbf{x})-\epsilon_{t,i},0\}\\
			&+\sum_j\max\{|h_j(\mathbf{x})|-\epsilon_{t,p+j},0\}.
		\end{aligned}
		\label{eq:e-violation}
	\end{equation}
	Larger entries in $\boldsymbol{\epsilon}_t$ let nearly feasible solutions support objective search, whereas smaller entries restore stricter feasibility pressure. MeCO learns when to relax and when to tighten this rule.
	
	Inside the SHADE optimizer, the relaxed violation is used only for candidate comparison. Given two candidates, if both have $\nu_{\boldsymbol{\epsilon}}=0$, the one with the smaller objective value wins. If only one has $\nu_{\boldsymbol{\epsilon}}=0$, the relaxed-feasible one wins. If neither is relaxed feasible, the one with the smaller $\nu_{\boldsymbol{\epsilon}}$ wins, and the objective is used only as a tie-breaker. Thus, the policy does not change mutation or crossover; it only changes how strongly the selection rule prefers feasibility over objective improvement at each step. Final test performance is still measured with the original unrelaxed violation.
	
	\subsection{Meta-Level MDP}
	\label{method:mdp}
	As shown in Fig.~\ref{fig:workflow}, the policy-optimizer interaction is modeled as
	$\mathcal{M}=\langle\mathcal{S},\mathcal{A},\mathcal{T},\mathcal{R},\pi_\theta\rangle$.
	At each step, $\mathcal{T}$ is induced by one SHADE update under the relaxation vector decoded from $\pi_\theta$'s action. For a training distribution $\mathcal{D}=\{\mathcal{D}_k\}_{k=1}^{F}$, the meta objective is
	\begin{equation}
		\label{eq:meta_objective}
		\theta^\star=\operatorname*{arg\,max}_{\theta}
		\frac{1}{F}\sum_{k=1}^{F}
		\mathbb{E}_{\pi_\theta,\mathcal{T}}
		\left[\sum_{t=1}^{T}\gamma^{t-1}r_t\mid\mathcal{D}_k\right].
	\end{equation}
	The objective encourages a policy that gives high accumulated optimization gain across training problems, rather than a policy tuned to one instance.
	At step $t$, the population $P_t$ is summarized as a 10-dimensional state vector
	\begin{equation}
		\mathbf{s}_t=[s_1,s_2,\ldots,s_{10}]^\top,
	\end{equation}
	with features
	\begin{equation}
		\label{eq:state_features}
		\begin{aligned}
			s_1&=\operatorname{std}\!\left(\frac{x_i-x^L}{x^U-x^L}\right),\\
			s_2&=\operatorname{std}\!\left(\frac{f(x_i)-f_{\mathrm{best}}^{(t)}}{f_{\mathrm{worst}}^{(0)}-f_{\mathrm{best}}^{(0)}}\right),\\
			s_3&=\operatorname{mean}\!\left(\frac{x_i-x^L}{x^U-x^L}\right),\\
			s_4&=\operatorname{mean}\!\left(\frac{f(x_i)-f_{\mathrm{best}}^{(t)}}{f_{\mathrm{worst}}^{(0)}-f_{\mathrm{best}}^{(0)}}\right),\\
			s_5&=\frac{f_{\mathrm{best}}^{(t)}}{f_{\mathrm{best}}^{(0)}},\quad
			s_6=\frac{\mathrm{FEs}}{\mathrm{FEs}_{\max}},\\
			s_7&=\frac{\bar{\nu}^{(t)}}{\bar{\nu}^{(0)}},\quad
			s_8=\frac{N_{\mathrm{feas}}}{N_{\mathrm{pop}}},\quad
			s_9=a_{t-1},\\
			s_{10}&=\frac{2}{N(N-1)}
			\sum_{i=1}^{N-1}\sum_{j=i+1}^{N}
			\mathbb{I}\!\left(\Delta f_{ij}\Delta\nu_{ij}>0\right),
		\end{aligned}
	\end{equation}
	where $\Delta f_{ij}=f(x_i)-f(x_j)$ and $\Delta\nu_{ij}=\nu(x_i)-\nu(x_j)$. The first six features summarize decision-space diversity, objective-space diversity, population centers, objective progress, and consumed evaluations. The last four features describe constraint-related information, including violation progress, feasibility ratio, the previous relaxation action, and the objective-constraint tradeoff. In the initial state, the previous action entry is set to 0. For numerical stability, denominators in the normalized features are bounded below by a small positive constant, and extreme normalized values are clipped in implementation. This compact design follows common designs in existing MetaBBOs~\cite{metabbosurvey,23ma2024auto23} and helps generalization across different problem and optimization settings.
	
	The action space is a coarse discrete grid $\mathcal{A}=\{0,0.1,\ldots,1\}$, which is used to keep the controller lightweight and reduce overfitting to a particular violation scale. The selected action is decoded into
	\begin{equation}
		\boldsymbol{\epsilon}_t=(\boldsymbol{\epsilon}_{\mathrm{base}})^{a_t}\cdot\delta^{1-a_t},
		\label{eq:epsilon}
	\end{equation}
	where the power operation is applied element by element, $\boldsymbol{\epsilon}_{\mathrm{base}}$ is the average initial violation vector clipped below by $\delta$, and $\delta=10^{-3}$. The lower clipping avoids a degenerate scale when some constraints are nearly satisfied in the initial population. Thus, $a_t=1$ gives the loosest comparison, $a_t=0$ gives the strictest comparison, and intermediate actions interpolate geometrically across violation scales. This scale-based interpolation gives the same discrete action a consistent meaning across problems with different constraint magnitudes.
	
	The reward has two normalized progress terms, objective improvement and violation reduction:
	\begin{equation}
		\begin{aligned}
			r_1^{(t)}&=
			\frac{f_{\mathrm{best}}^{(t-1)}-f_{\mathrm{best}}^{(t)}}{f_{\mathrm{best}}^{(0)}-\tilde{f}^\star},\\
			r_2^{(t)}&=
			\frac{\bar{\nu}^{(t-1)}-\bar{\nu}^{(t)}}{\bar{\nu}^{(0)}}.
		\end{aligned}
		\label{eq:reward_components}
	\end{equation}
	The full reward is defined as
	\begin{equation}
		r_t=\frac{(1-\omega_t)r_1^{(t)}+r_2^{(t)}}{2},
		\quad
		\omega_t=\frac{\bar{\nu}^{(t)}}{\bar{\nu}^{(0)}} .
		\label{eq:reward}
	\end{equation}
	
	
	Here \(\bar{\nu}^{(t)}\) is the average of the five smallest violations in the current population, and \(\tilde{f}^{\star}\) is a proxy reference objective value used only for reward normalization. In our implementation, \(\tilde{f}^{\star}\) is set to the best objective value found by the agent during meta-training for the corresponding training problem. When violations remain large, objective improvement receives less weight; as feasibility improves, the reward gives more weight to objective refinement.
	
	MeCO uses a Double Deep Q-Network (DDQN)~\cite{25van2016deep25} with one hidden layer of 64 SELU units and 11 outputs. During training, each episode is a complete optimization run on a sampled constrained problem. The transition $(\mathbf{s}_t,a_t,r_t,\mathbf{s}_{t+1})$ is stored in replay memory. With online parameters $\theta$ and target parameters $\theta'$, the update target is
	\begin{equation}
		\label{eq:ddqn_loss}
		\begin{aligned}
			a^\star&=\operatorname*{arg\,max}_{a'}Q(\mathbf{s}_{t+1},a';\theta),\\
			y_t&=r_t+\gamma Q(\mathbf{s}_{t+1},a^\star;\theta'),\\
			\mathcal{L}(\theta)&=\left(y_t-Q(\mathbf{s}_t,a_t;\theta)\right)^2.
		\end{aligned}
	\end{equation}
	The target network is copied from the online network periodically. After training, the policy is fixed and directly controls SHADE on held-out instances.
	
	\section{Experiments}
	\label{sec:exp}
	
	\subsection{Setup}
	We evaluate MeCO on the CEC2017 constrained real-parameter benchmark~\cite{cec2017}, which contains 28 shifted, rotated, separable, non-separable, narrow-feasible, and boundary-constrained problems. Unless otherwise stated, each method is run 10 times under a low-budget setting with $\mathrm{FEs}_{\max}=50D$. This protocol tests online decision quality under limited evaluations and keeps the benchmark budget comparable to the UAV transfer task, which uses 2500 evaluations.
	
	We compare MeCO with ten external baselines organized according to their constraint-handling mechanisms: (I) penalty- or adaptive-penalty-based methods, including \textbf{SHADE}~\cite{tanabe2013success}, which uses a static penalty with $\lambda=1$, \textbf{SHADE-BOC}, which applies the adaptive penalty rule of L-SHADE-BOC~\cite{kawachi2019shade}, and \textbf{CMA-ES}~\cite{cmaes}, which uses the same static penalty as SHADE; (II) relaxation-based methods, including \textbf{SHADE-$\epsilon$}, which uses a fixed relaxation level equal to $0.1$ times the maximum initial violation, \textbf{MDE-CGO}~\cite{bai2025uav}, which prioritizes constraint violation during selection, \textbf{C$^2$oDE}~\cite{c2ode}, which combines violation-based selection with $\epsilon$-relaxation, and \textbf{BP-$\epsilon$MAg-ES}~\cite{9185566}, which uses $\epsilon$-level selection within a restart-based matrix-adaptation ES; (III) surrogate-assisted methods, represented by \textbf{COBRA}~\cite{6regis2014constrained6} and \textbf{SACOBRA}~\cite{sacobra}; and (IV) multi-objective methods, represented by \textbf{NSGA-III}~\cite{deb2013evolutionary}, which treats the objective and constraints in a multi-objective formulation. We also include \textbf{MeCO-u}, an untrained MeCO variant with randomly initialized DDQN parameters. Table~\ref{tab:comparison} reports $f(\mathbf{x}^\star)+\nu(\mathbf{x}^\star)$, where $\nu$ is the original unrelaxed violation, equivalently Eq.~\eqref{eq:e-violation} with $\boldsymbol{\epsilon}=\mathbf{0}$. Lower values are better.
	
	For fairness, all methods use the same evaluation budget, stopping rule, box constraints, final violation definition, and final metric. Population-based methods use a population size of 50; surrogate-assisted methods follow their recommended settings under the same budget. Baseline parameters follow the original papers unless alignment is required by the shared low-budget protocol.
	
	During meta-training, MeCO uses a replay memory capacity of 64, a mini-batch size of 8, and 50 training epochs. AdamW uses a linearly decayed learning rate from $5\times10^{-3}$ to $10^{-4}$; $\epsilon$-greedy exploration is 0.1, the target Q-network is updated every 10 learning steps, and $\gamma=1$. The proxy reference value \(\tilde{f}^{\star}\) is set to the best objective value found by the agent during meta-training for the corresponding training problem.
	
	\begin{table}[!t]
		\centering
		\caption{Leave one out generalization on CEC2017 with $D=50$. Each entry reports $f(\mathbf{x}^\star)+\nu(\mathbf{x}^\star)$, where $\nu$ is the original unrelaxed constraint violation. Lower values are better.}
		\label{tab:comparison}
		
		\setlength{\tabcolsep}{1.5pt}
		\renewcommand{\arraystretch}{0.9}
		
		\resizebox{\linewidth}{!}{
			\begin{tabular}{
					>{\centering\arraybackslash}m{0.65cm}
					*{12}{>{\centering\arraybackslash}m{1.85cm}}
				}
				\toprule
				
				$F$
				& SHADE-$\epsilon$
				& SHADE~\cite{tanabe2013success}
				& SHADE-BOC~\cite{kawachi2019shade}
				& MDE-CGO~\cite{bai2025uav}
				& C$^2$oDE~\cite{c2ode}
				& COBRA~\cite{6regis2014constrained6}
				& SACOBRA~\cite{sacobra}
				& NSGA-III~\cite{deb2013evolutionary}
				& BP-$\epsilon$MAg-ES~\cite{9185566}
				& CMA-ES~\cite{cmaes}
				& MeCO-u
				& MeCO
				\\
				
				\midrule
				
				$F_{1}$ &
				\makecell{2.74e+05 \\ $\pm$5.29e+04} &
				\makecell{1.27e+05 \\ $\pm$4.15e+04} &
				\underline{\makecell{1.09e+05 \\ $\pm$5.37e+04}} &
				\makecell{2.61e+05 \\ $\pm$1.76e+04} &
				\makecell{1.64e+05 \\ $\pm$3.96e+04} &
				\makecell{3.41e+05 \\ $\pm$5.53e+04} &
				\makecell{1.28e+05 \\ $\pm$4.66e+04} &
				\makecell{2.46e+05 \\ $\pm$4.30e+04} &
				\makecell{1.39e+05 \\ $\pm$5.17e+04} &
				\makecell{1.17e+05 \\ $\pm$3.16e+04} &
				\makecell{2.40e+05 \\ $\pm$3.16e+04} &
				\textbf{\makecell{1.01e+05 \\ $\pm$4.54e+04}} \\
				$F_{2}$ &
				\makecell{2.68e+05 \\ $\pm$6.96e+04} &
				\makecell{1.26e+05 \\ $\pm$5.00e+04} &
				\makecell{1.19e+05 \\ $\pm$4.05e+04} &
				\makecell{2.82e+05 \\ $\pm$5.49e+04} &
				\makecell{1.96e+05 \\ $\pm$6.05e+04} &
				\makecell{3.89e+05 \\ $\pm$8.49e+04} &
				\makecell{1.25e+05 \\ $\pm$4.44e+04} &
				\makecell{1.08e+05 \\ $\pm$4.85e+04} &
				\makecell{1.33e+05 \\ $\pm$4.37e+04} &
				\underline{\makecell{9.55e+04 \\ $\pm$3.71e+04}} &
				\makecell{2.10e+05 \\ $\pm$3.06e+04} &
				\textbf{\makecell{9.01e+04 \\ $\pm$2.61e+04}} \\
				$F_{3}$ &
				\makecell{1.12e+05 \\ $\pm$2.49e+04} &
				\makecell{2.69e+05 \\ $\pm$9.45e+04} &
				\makecell{2.17e+05 \\ $\pm$6.78e+04} &
				\makecell{1.13e+05 \\ $\pm$2.39e+04} &
				\textbf{\makecell{6.73e+04 \\ $\pm$1.39e+04}} &
				\makecell{1.46e+05 \\ $\pm$1.32e+04} &
				\makecell{1.34e+05 \\ $\pm$3.99e+04} &
				\makecell{1.54e+05 \\ $\pm$3.44e+04} &
				\makecell{3.52e+05 \\ $\pm$1.86e+05} &
				\underline{\makecell{7.73e+04 \\ $\pm$1.18e+04}} &
				\makecell{2.12e+05 \\ $\pm$6.23e+04} &
				\makecell{1.86e+05 \\ $\pm$4.69e+04} \\
				$F_{4}$ &
				\makecell{1.72e+03 \\ $\pm$1.06e+02} &
				\makecell{1.14e+03 \\ $\pm$1.99e+02} &
				\makecell{1.42e+03 \\ $\pm$2.98e+02} &
				\makecell{1.04e+03 \\ $\pm$5.46e+01} &
				\makecell{1.75e+03 \\ $\pm$1.79e+02} &
				\makecell{2.13e+03 \\ $\pm$1.25e+02} &
				\makecell{1.97e+03 \\ $\pm$1.96e+02} &
				\makecell{1.40e+03 \\ $\pm$1.23e+02} &
				\makecell{1.01e+03 \\ $\pm$5.52e+02} &
				\textbf{\makecell{5.79e+02 \\ $\pm$2.00e+01}} &
				\makecell{1.76e+03 \\ $\pm$1.93e+02} &
				\underline{\makecell{9.52e+02 \\ $\pm$1.85e+02}} \\
				$F_{5}$ &
				\makecell{4.65e+06 \\ $\pm$1.19e+06} &
				\makecell{1.67e+06 \\ $\pm$1.80e+06} &
				\makecell{1.41e+06 \\ $\pm$1.07e+06} &
				\makecell{1.92e+06 \\ $\pm$2.85e+05} &
				\makecell{2.93e+06 \\ $\pm$7.99e+05} &
				\makecell{1.19e+07 \\ $\pm$1.24e+06} &
				\makecell{1.22e+07 \\ $\pm$1.90e+06} &
				\makecell{5.26e+06 \\ $\pm$8.77e+05} &
				\underline{\makecell{5.23e+05 \\ $\pm$7.84e+05}} &
				\textbf{\makecell{4.85e+04 \\ $\pm$1.15e+04}} &
				\makecell{5.67e+06 \\ $\pm$2.36e+06} &
				\makecell{7.60e+05 \\ $\pm$5.72e+05} \\
				$F_{6}$ &
				\makecell{8.16e+03 \\ $\pm$8.17e+02} &
				\makecell{8.94e+03 \\ $\pm$1.29e+03} &
				\makecell{9.38e+03 \\ $\pm$1.74e+03} &
				\underline{\makecell{3.45e+03 \\ $\pm$7.02e+01}} &
				\makecell{8.08e+03 \\ $\pm$8.11e+02} &
				\makecell{8.60e+03 \\ $\pm$6.00e+02} &
				\makecell{8.41e+03 \\ $\pm$3.28e+02} &
				\makecell{4.06e+03 \\ $\pm$2.64e+02} &
				\makecell{4.84e+03 \\ $\pm$1.40e+03} &
				\makecell{8.77e+03 \\ $\pm$4.20e+01} &
				\makecell{7.11e+03 \\ $\pm$8.53e+02} &
				\textbf{\makecell{2.38e+03 \\ $\pm$1.14e+03}} \\
				$F_{7}$ &
				\makecell{6.37e+03 \\ $\pm$4.32e+02} &
				\makecell{4.93e+03 \\ $\pm$6.40e+02} &
				\makecell{5.46e+03 \\ $\pm$4.57e+02} &
				\makecell{6.70e+03 \\ $\pm$2.56e+02} &
				\makecell{4.60e+03 \\ $\pm$2.89e+02} &
				\makecell{5.55e+03 \\ $\pm$3.08e+02} &
				\makecell{6.18e+03 \\ $\pm$3.97e+02} &
				\underline{\makecell{4.41e+03 \\ $\pm$5.07e+02}} &
				\textbf{\makecell{1.35e+02 \\ $\pm$2.19e+02}} &
				\makecell{5.20e+03 \\ $\pm$2.25e+02} &
				\makecell{5.49e+03 \\ $\pm$4.90e+02} &
				\makecell{4.66e+03 \\ $\pm$4.77e+02} \\
				$F_{8}$ &
				\makecell{3.11e+05 \\ $\pm$6.44e+04} &
				\makecell{1.73e+05 \\ $\pm$5.11e+04} &
				\makecell{1.43e+05 \\ $\pm$3.72e+04} &
				\makecell{2.48e+05 \\ $\pm$3.56e+04} &
				\makecell{1.53e+05 \\ $\pm$3.46e+04} &
				\makecell{2.76e+05 \\ $\pm$5.55e+04} &
				\makecell{1.68e+05 \\ $\pm$1.12e+05} &
				\underline{\makecell{1.32e+05 \\ $\pm$3.48e+04}} &
				\makecell{2.68e+05 \\ $\pm$1.03e+05} &
				\makecell{2.10e+05 \\ $\pm$4.26e+04} &
				\makecell{3.46e+05 \\ $\pm$3.80e+04} &
				\textbf{\makecell{1.18e+05 \\ $\pm$3.01e+04}} \\
				$F_{9}$ &
				\makecell{9.92e+04 \\ $\pm$4.37e+04} &
				\makecell{5.23e+04 \\ $\pm$4.24e+04} &
				\makecell{3.65e+04 \\ $\pm$2.37e+04} &
				\makecell{4.50e+04 \\ $\pm$1.17e+04} &
				\makecell{1.51e+05 \\ $\pm$9.47e+04} &
				\makecell{6.85e+04 \\ $\pm$1.59e+04} &
				\makecell{2.38e+04 \\ $\pm$1.65e+04} &
				\makecell{6.41e+04 \\ $\pm$1.97e+04} &
				\underline{\makecell{1.87e+04 \\ $\pm$7.62e+03}} &
				\makecell{1.92e+05 \\ $\pm$2.50e+04} &
				\makecell{2.97e+04 \\ $\pm$1.50e+04} &
				\textbf{\makecell{1.71e+04 \\ $\pm$6.10e+03}} \\
				$F_{10}$ &
				\makecell{7.54e+05 \\ $\pm$6.31e+04} &
				\makecell{3.44e+05 \\ $\pm$1.17e+05} &
				\makecell{4.03e+05 \\ $\pm$7.49e+04} &
				\makecell{5.86e+05 \\ $\pm$2.86e+04} &
				\makecell{5.00e+05 \\ $\pm$6.24e+04} &
				\makecell{8.56e+05 \\ $\pm$7.64e+04} &
				\makecell{3.67e+05 \\ $\pm$1.87e+05} &
				\makecell{6.06e+05 \\ $\pm$1.03e+05} &
				\textbf{\makecell{1.13e+05 \\ $\pm$9.29e+04}} &
				\underline{\makecell{2.21e+05 \\ $\pm$3.73e+04}} &
				\makecell{4.71e+05 \\ $\pm$8.03e+04} &
				\makecell{3.67e+05 \\ $\pm$1.47e+05} \\
				$F_{11}$ &
				\makecell{3.32e+05 \\ $\pm$2.83e+04} &
				\makecell{7.72e+04 \\ $\pm$3.84e+04} &
				\makecell{1.67e+05 \\ $\pm$7.62e+04} &
				\makecell{3.24e+05 \\ $\pm$3.39e+04} &
				\makecell{3.02e+05 \\ $\pm$2.91e+04} &
				\makecell{2.62e+05 \\ $\pm$1.76e+04} &
				\makecell{2.10e+05 \\ $\pm$6.54e+04} &
				\makecell{2.16e+05 \\ $\pm$3.01e+04} &
				\textbf{\makecell{5.12e+04 \\ $\pm$6.74e+04}} &
				\underline{\makecell{5.38e+04 \\ $\pm$1.20e+04}} &
				\makecell{1.95e+05 \\ $\pm$1.92e+04} &
				\makecell{9.09e+04 \\ $\pm$1.68e+04} \\
				$F_{12}$ &
				\makecell{2.00e+05 \\ $\pm$1.11e+04} &
				\makecell{7.42e+04 \\ $\pm$3.98e+04} &
				\makecell{7.21e+04 \\ $\pm$2.88e+04} &
				\makecell{1.12e+05 \\ $\pm$1.13e+04} &
				\makecell{1.12e+05 \\ $\pm$2.28e+04} &
				\makecell{2.85e+05 \\ $\pm$2.03e+04} &
				\makecell{2.05e+05 \\ $\pm$1.64e+04} &
				\makecell{1.72e+05 \\ $\pm$2.38e+04} &
				\textbf{\makecell{1.84e+03 \\ $\pm$2.34e+03}} &
				\underline{\makecell{1.51e+04 \\ $\pm$2.44e+03}} &
				\makecell{1.19e+05 \\ $\pm$3.41e+04} &
				\makecell{6.38e+04 \\ $\pm$3.09e+04} \\
				$F_{13}$ &
				\makecell{4.62e+10 \\ $\pm$8.88e+09} &
				\makecell{1.70e+10 \\ $\pm$1.21e+10} &
				\makecell{1.46e+10 \\ $\pm$1.22e+10} &
				\makecell{2.06e+10 \\ $\pm$6.02e+09} &
				\makecell{2.28e+10 \\ $\pm$6.10e+09} &
				\makecell{1.07e+11 \\ $\pm$1.71e+10} &
				\makecell{7.78e+10 \\ $\pm$1.31e+10} &
				\makecell{5.46e+10 \\ $\pm$1.14e+10} &
				\underline{\makecell{3.60e+08 \\ $\pm$4.45e+08}} &
				\textbf{\makecell{2.26e+08 \\ $\pm$1.04e+08}} &
				\makecell{1.92e+10 \\ $\pm$7.43e+09} &
				\makecell{9.83e+09 \\ $\pm$8.15e+09} \\
				$F_{14}$ &
				\makecell{1.89e+05 \\ $\pm$1.44e+04} &
				\makecell{9.92e+04 \\ $\pm$4.93e+04} &
				\makecell{9.30e+04 \\ $\pm$3.71e+04} &
				\makecell{3.97e+05 \\ $\pm$2.91e+04} &
				\makecell{1.39e+05 \\ $\pm$2.23e+04} &
				\makecell{2.81e+05 \\ $\pm$2.02e+04} &
				\makecell{1.27e+05 \\ $\pm$3.98e+04} &
				\makecell{2.33e+05 \\ $\pm$1.78e+04} &
				\textbf{\makecell{1.63e+03 \\ $\pm$2.73e+04}} &
				\makecell{8.58e+04 \\ $\pm$2.53e+04} &
				\makecell{1.44e+05 \\ $\pm$5.47e+04} &
				\underline{\makecell{7.97e+04 \\ $\pm$4.28e+04}} \\
				$F_{15}$ &
				\makecell{9.16e+04 \\ $\pm$8.00e+03} &
				\makecell{4.60e+04 \\ $\pm$1.94e+04} &
				\makecell{3.80e+04 \\ $\pm$1.88e+04} &
				\makecell{9.15e+04 \\ $\pm$9.79e+03} &
				\makecell{5.97e+04 \\ $\pm$9.52e+03} &
				\makecell{1.38e+05 \\ $\pm$1.15e+04} &
				\makecell{8.69e+04 \\ $\pm$1.67e+04} &
				\makecell{1.21e+05 \\ $\pm$1.50e+04} &
				\underline{\makecell{1.50e+04 \\ $\pm$7.17e+03}} &
				\textbf{\makecell{3.83e+03 \\ $\pm$1.18e+03}} &
				\makecell{2.16e+05 \\ $\pm$1.76e+04} &
				\makecell{3.38e+04 \\ $\pm$2.24e+04} \\
				$F_{16}$ &
				\makecell{8.94e+04 \\ $\pm$5.75e+03} &
				\makecell{2.95e+04 \\ $\pm$1.54e+04} &
				\makecell{3.75e+04 \\ $\pm$2.84e+04} &
				\makecell{5.12e+04 \\ $\pm$4.46e+03} &
				\makecell{6.51e+04 \\ $\pm$4.24e+03} &
				\makecell{1.37e+05 \\ $\pm$1.42e+04} &
				\makecell{3.16e+04 \\ $\pm$1.36e+04} &
				\makecell{9.98e+04 \\ $\pm$9.69e+03} &
				\makecell{3.51e+04 \\ $\pm$2.46e+04} &
				\textbf{\makecell{2.07e+04 \\ $\pm$1.04e+04}} &
				\makecell{7.95e+04 \\ $\pm$8.81e+03} &
				\underline{\makecell{2.69e+04 \\ $\pm$1.78e+04}} \\
				$F_{17}$ &
				\makecell{9.93e+04 \\ $\pm$1.35e+04} &
				\makecell{5.14e+04 \\ $\pm$2.61e+04} &
				\underline{\makecell{4.06e+04 \\ $\pm$1.72e+04}} &
				\makecell{5.29e+04 \\ $\pm$2.06e+03} &
				\makecell{6.34e+04 \\ $\pm$1.32e+04} &
				\makecell{1.35e+05 \\ $\pm$1.28e+04} &
				\makecell{1.92e+05 \\ $\pm$2.30e+04} &
				\makecell{9.22e+04 \\ $\pm$1.49e+04} &
				\makecell{5.62e+04 \\ $\pm$1.24e+04} &
				\makecell{7.34e+04 \\ $\pm$9.89e+03} &
				\makecell{4.98e+04 \\ $\pm$1.35e+04} &
				\textbf{\makecell{3.93e+04 \\ $\pm$1.81e+04}} \\
				$F_{18}$ &
				\makecell{4.86e+10 \\ $\pm$1.10e+10} &
				\makecell{1.01e+10 \\ $\pm$1.08e+10} &
				\makecell{1.08e+10 \\ $\pm$9.42e+09} &
				\makecell{1.34e+10 \\ $\pm$3.18e+09} &
				\makecell{3.33e+10 \\ $\pm$1.07e+10} &
				\makecell{1.17e+11 \\ $\pm$1.09e+10} &
				\textbf{\makecell{3.81e+08 \\ $\pm$1.07e+08}} &
				\makecell{4.64e+10 \\ $\pm$7.05e+09} &
				\makecell{6.18e+09 \\ $\pm$7.27e+09} &
				\underline{\makecell{3.82e+08 \\ $\pm$9.97e+08}} &
				\makecell{6.23e+10 \\ $\pm$6.78e+09} &
				\makecell{9.16e+09 \\ $\pm$4.18e+09} \\
				$F_{19}$ &
				\makecell{7.25e+04 \\ $\pm$4.18e+01} &
				\makecell{7.22e+04 \\ $\pm$1.01e+02} &
				\makecell{7.22e+04 \\ $\pm$8.85e+01} &
				\makecell{7.22e+04 \\ $\pm$9.15e+01} &
				\makecell{7.23e+04 \\ $\pm$6.39e+01} &
				\makecell{7.24e+04 \\ $\pm$6.47e+01} &
				\makecell{7.28e+04 \\ $\pm$2.78e+01} &
				\underline{\makecell{7.21e+04 \\ $\pm$1.33e+02}} &
				\makecell{7.29e+04 \\ $\pm$5.79e+01} &
				\makecell{7.28e+04 \\ $\pm$1.26e+01} &
				\makecell{7.24e+04 \\ $\pm$7.05e+01} &
				\textbf{\makecell{7.21e+04 \\ $\pm$7.53e+01}} \\
				$F_{20}$ &
				\makecell{1.94e+01 \\ $\pm$8.28e-01} &
				\makecell{1.84e+01 \\ $\pm$5.80e-01} &
				\makecell{1.85e+01 \\ $\pm$1.16e+00} &
				\underline{\makecell{1.75e+01 \\ $\pm$5.32e-01}} &
				\makecell{1.75e+01 \\ $\pm$6.52e-01} &
				\makecell{1.80e+01 \\ $\pm$9.55e-01} &
				\makecell{1.88e+01 \\ $\pm$7.22e-01} &
				\makecell{1.77e+01 \\ $\pm$5.93e-01} &
				\makecell{1.91e+01 \\ $\pm$5.85e-01} &
				\makecell{1.84e+01 \\ $\pm$5.48e-01} &
				\makecell{1.89e+01 \\ $\pm$7.28e-01} &
				\textbf{\makecell{1.74e+01 \\ $\pm$1.18e+00}} \\
				$F_{21}$ &
				\makecell{1.92e+05 \\ $\pm$1.90e+04} &
				\makecell{7.42e+04 \\ $\pm$3.98e+04} &
				\underline{\makecell{7.21e+04 \\ $\pm$2.88e+04}} &
				\makecell{1.09e+05 \\ $\pm$1.46e+04} &
				\makecell{1.18e+05 \\ $\pm$2.09e+04} &
				\makecell{2.84e+05 \\ $\pm$1.87e+04} &
				\makecell{1.79e+05 \\ $\pm$2.68e+04} &
				\makecell{1.71e+05 \\ $\pm$2.58e+04} &
				\makecell{1.80e+05 \\ $\pm$2.28e+04} &
				\makecell{8.48e+04 \\ $\pm$2.21e+04} &
				\makecell{9.22e+04 \\ $\pm$3.12e+04} &
				\textbf{\makecell{6.90e+04 \\ $\pm$3.40e+04}} \\
				$F_{22}$ &
				\makecell{4.53e+10 \\ $\pm$1.21e+10} &
				\makecell{8.47e+09 \\ $\pm$8.41e+09} &
				\makecell{1.60e+10 \\ $\pm$1.35e+10} &
				\makecell{2.13e+10 \\ $\pm$5.77e+09} &
				\makecell{2.52e+10 \\ $\pm$9.87e+09} &
				\makecell{1.04e+11 \\ $\pm$1.51e+10} &
				\makecell{1.03e+10 \\ $\pm$4.87e+09} &
				\makecell{4.44e+10 \\ $\pm$1.12e+10} &
				\makecell{4.06e+10 \\ $\pm$5.13e+09} &
				\textbf{\makecell{2.59e+09 \\ $\pm$7.72e+09}} &
				\makecell{1.94e+10 \\ $\pm$9.34e+10} &
				\underline{\makecell{4.82e+09 \\ $\pm$1.87e+09}} \\
				$F_{23}$ &
				\makecell{2.00e+05 \\ $\pm$2.01e+04} &
				\underline{\makecell{7.45e+04 \\ $\pm$4.70e+04}} &
				\makecell{1.03e+05 \\ $\pm$4.51e+04} &
				\makecell{4.26e+05 \\ $\pm$2.65e+04} &
				\makecell{1.32e+05 \\ $\pm$2.22e+04} &
				\makecell{2.74e+05 \\ $\pm$2.72e+04} &
				\makecell{1.80e+05 \\ $\pm$6.63e+04} &
				\makecell{2.34e+05 \\ $\pm$3.30e+04} &
				\makecell{1.57e+05 \\ $\pm$2.69e+04} &
				\makecell{1.70e+05 \\ $\pm$2.50e+04} &
				\makecell{9.38e+04 \\ $\pm$3.59e+04} &
				\textbf{\makecell{6.87e+04 \\ $\pm$3.84e+04}} \\
				$F_{24}$ &
				\makecell{8.90e+04 \\ $\pm$9.32e+03} &
				\makecell{4.26e+04 \\ $\pm$2.98e+04} &
				\makecell{3.86e+04 \\ $\pm$2.46e+04} &
				\makecell{1.07e+05 \\ $\pm$1.43e+04} &
				\makecell{6.05e+04 \\ $\pm$1.00e+04} &
				\makecell{1.34e+05 \\ $\pm$4.53e+03} &
				\makecell{9.51e+04 \\ $\pm$7.93e+03} &
				\makecell{1.10e+05 \\ $\pm$1.71e+04} &
				\underline{\makecell{2.37e+03 \\ $\pm$2.55e+03}} &
				\textbf{\makecell{2.21e+03 \\ $\pm$1.53e+03}} &
				\makecell{4.15e+04 \\ $\pm$2.21e+04} &
				\makecell{2.80e+04 \\ $\pm$2.41e+04} \\
				$F_{25}$ &
				\makecell{9.09e+04 \\ $\pm$1.22e+04} &
				\makecell{5.34e+04 \\ $\pm$1.80e+04} &
				\makecell{5.27e+04 \\ $\pm$3.48e+04} &
				\makecell{5.96e+04 \\ $\pm$7.37e+03} &
				\makecell{5.93e+04 \\ $\pm$1.13e+04} &
				\makecell{1.42e+05 \\ $\pm$9.61e+03} &
				\makecell{6.63e+04 \\ $\pm$7.40e+03} &
				\makecell{4.04e+04 \\ $\pm$1.32e+04} &
				\makecell{4.35e+04 \\ $\pm$1.86e+04} &
				\underline{\makecell{3.72e+04 \\ $\pm$1.27e+04}} &
				\makecell{2.18e+05 \\ $\pm$2.04e+04} &
				\textbf{\makecell{3.69e+04 \\ $\pm$1.72e+04}} \\
				$F_{26}$ &
				\makecell{9.35e+04 \\ $\pm$1.14e+04} &
				\makecell{5.14e+04 \\ $\pm$2.61e+04} &
				\underline{\makecell{4.06e+04 \\ $\pm$1.72e+04}} &
				\makecell{5.58e+04 \\ $\pm$5.51e+03} &
				\makecell{6.21e+04 \\ $\pm$1.25e+04} &
				\makecell{1.37e+05 \\ $\pm$1.03e+04} &
				\makecell{2.00e+05 \\ $\pm$2.29e+04} &
				\makecell{8.85e+04 \\ $\pm$8.88e+03} &
				\makecell{8.51e+04 \\ $\pm$1.38e+03} &
				\makecell{7.52e+05 \\ $\pm$1.34e+04} &
				\makecell{5.02e+04 \\ $\pm$1.95e+04} &
				\textbf{\makecell{3.94e+04 \\ $\pm$1.78e+04}} \\
				$F_{27}$ &
				\makecell{4.50e+10 \\ $\pm$6.94e+09} &
				\underline{\makecell{6.93e+09 \\ $\pm$6.73e+09}} &
				\makecell{1.37e+10 \\ $\pm$1.21e+10} &
				\makecell{1.49e+10 \\ $\pm$2.13e+09} &
				\makecell{3.15e+10 \\ $\pm$5.86e+09} &
				\makecell{1.16e+11 \\ $\pm$1.73e+10} &
				\makecell{4.50e+10 \\ $\pm$1.82e+10} &
				\makecell{4.48e+10 \\ $\pm$1.05e+10} &
				\textbf{\makecell{2.59e+09 \\ $\pm$3.68e+09}} &
				\makecell{4.63e+10 \\ $\pm$5.50e+09} &
				\makecell{7.16e+09 \\ $\pm$3.52e+09} &
				\makecell{7.01e+09 \\ $\pm$4.19e+09} \\
				$F_{28}$ &
				\makecell{7.25e+04 \\ $\pm$3.24e+01} &
				\makecell{7.24e+04 \\ $\pm$7.85e+01} &
				\underline{\makecell{7.23e+04 \\ $\pm$1.51e+02}} &
				\makecell{7.24e+04 \\ $\pm$6.16e+01} &
				\makecell{7.25e+04 \\ $\pm$3.81e+01} &
				\makecell{7.25e+04 \\ $\pm$6.06e+01} &
				\makecell{7.28e+04 \\ $\pm$2.86e+01} &
				\makecell{7.25e+04 \\ $\pm$4.70e+01} &
				\makecell{7.28e+04 \\ $\pm$6.12e+01} &
				\makecell{7.27e+04 \\ $\pm$9.38e+01} &
				\makecell{7.25e+04 \\ $\pm$1.32e+02} &
				\textbf{\makecell{7.23e+04 \\ $\pm$1.24e+02}} \\
				\midrule
				Avg Rank & 9.36 & 5.00 & 4.82 & 7.07 & 6.68 & 10.43 & 7.82 & 7.36 & 5.07 & \underline{4.64} & 7.50 & \textbf{2.25} \\
				\bottomrule
			\end{tabular}
		}
	\end{table}

	\begin{figure}[!t]
		\centering
		\includegraphics[width=0.70\linewidth]{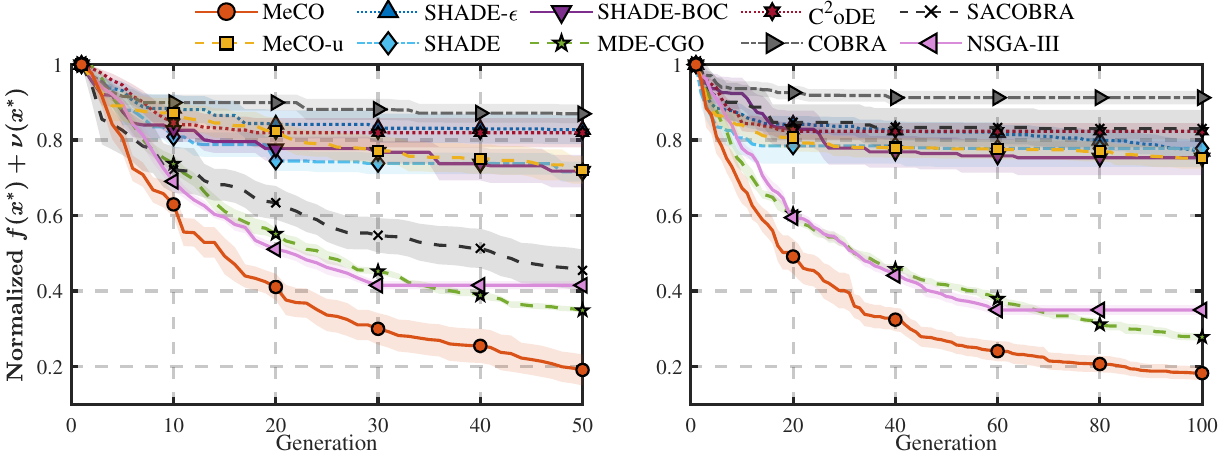}
		\caption{Dimension transfer on 50D and 100D CEC2017 problems. MeCO is trained only on 10D problems. Curves report normalized objective-plus-violation values under the original unrelaxed violation, where lower values are better.}
		\label{fig:wdfh}
	\end{figure}
	
	\subsection{Generalization Performance}
	
	Table~\ref{tab:comparison} reports leave-one-out results on 50D CEC2017, where each MeCO model is trained on 27 functions and tested on the held-out function. MeCO achieves the best average rank, indicating strong overall generalization across heterogeneous constrained functions. Several baselines obtain the best value on individual functions, showing that specialized constraint-handling mechanisms can be highly competitive on specific landscapes. In contrast, MeCO is not tied to a fixed penalty, relaxation level, or feasibility rule; it learns a state-dependent relaxation policy that remains consistently competitive across penalty-sensitive, relaxation-sensitive, and large-violation cases. The comparisons with SHADE-$\epsilon$, SHADE-BOC, and MeCO-u further indicate that the learned DDQN policy, rather than a fixed relaxation level, an adaptive penalty rule, or a random controller, is the key source of adaptive feasibility pressure.
	
	Dimension transfer tests whether the controller acts on normalized search states rather than a fixed dimension. The policy is trained on 10D problems and then used directly on 50D and 100D problems. Fig.~\ref{fig:wdfh} shows that MeCO maintains better normalized progress in both settings. Handcrafted baselines often improve quickly and then plateau, while MeCO can relax the rule when nearly feasible candidates are useful and tighten it when feasibility becomes the bottleneck.

	\begin{figure}[!t]
		\centering
		\includegraphics[width=0.80\linewidth]{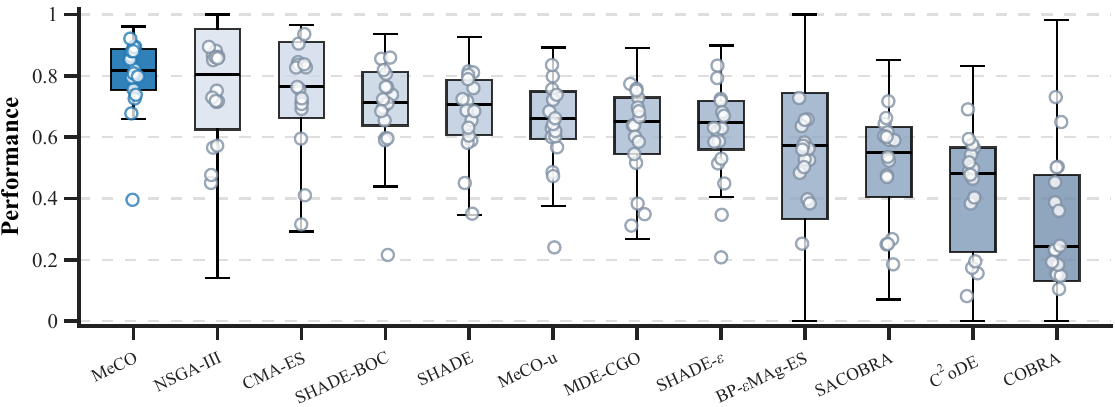}
		\caption{UAV path-planning transfer experiment. The boxplot reports normalized performance on the UAV tasks. Higher values indicate better performance.}
		\label{fig:uavsss}
	\end{figure}
	
	\subsection{Application-Domain Transfer}
	We further evaluate application-domain transfer on 16 UAV path-planning tasks for wireless-sensor data collection~\cite{bai2025uav} and eight real-world constrained engineering design problems~\cite{kumar2020test}, with a budget of 2500 function evaluations. In all transfer experiments, the MeCO policy is trained on 10D CEC2017 problems and is then directly deployed to the target tasks. Fig.~\ref{fig:uavsss} presents the results on the UAV tasks, each of which contains obstacle constraints and six hovering points. MeCO achieves strong normalized performance across these tasks, demonstrating its ability to transfer the learned constraint-handling strategy beyond the CEC2017 benchmark family. Results on the eight real-world engineering design problems are provided in the Supplementary Material.

	\subsection{Behavior and Ablation Evidence}
	
	\begin{figure}[!t]
		\centering
		\includegraphics[width=0.62\linewidth]{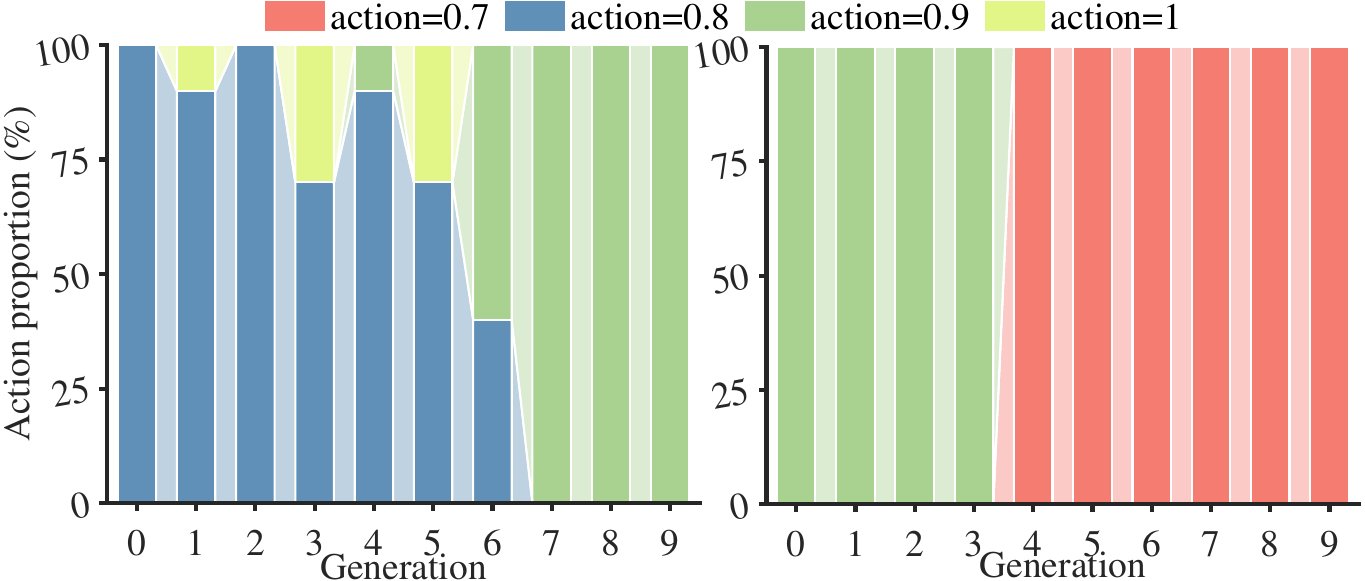}
		\caption{Selected $\epsilon$-relaxation actions on two CEC2017 functions. Left: $F_{12}$; right: $F_{14}$. Larger actions correspond to looser relaxation vectors.}
		\label{fig:actions}
	\end{figure}
	
	Fig.~\ref{fig:actions} shows that MeCO selects different relaxation actions for different feasibility structures. On $F_{12}$, the equality constraint forms a connected hyperspherical manifold, so nearly feasible solutions can guide objective improvement and the policy often selects looser actions. On $F_{14}$, feasibility depends on $\cos(f(\mathbf{x}))+\sin(f(\mathbf{x}))=0$, creating separated feasible shells tied to quantized objective values; the policy therefore shifts toward stricter actions. Thus, the controller adapts the feasibility-optimality tradeoff to the observed state rather than changing $\boldsymbol{\epsilon}$ monotonically.
	
	The component ablations use a fixed CEC2017 split: $F_7$, $F_8$, $F_{10}$, $F_{11}$, $F_{14}$, $F_{23}$, $F_{24}$, and $F_{28}$ are held out, while the remaining functions are used for training. This isolates each design choice on consistent held-out problems.
	
	\begin{figure}[!t]
		\centering
		\includegraphics[width=0.65\linewidth]{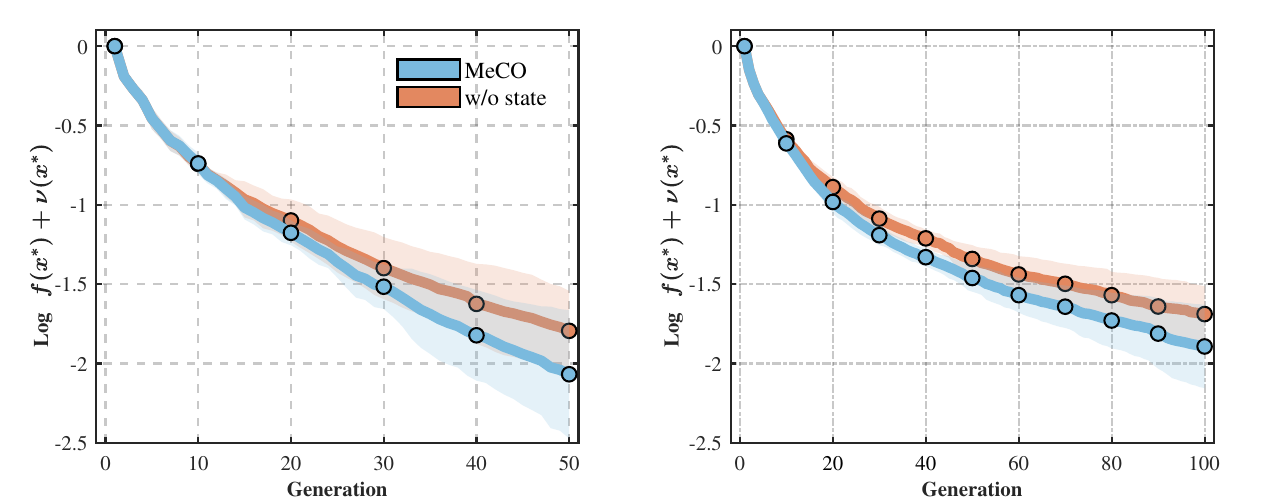}
		\caption{State-feature ablation on the fixed CEC2017 test set for 50D and 100D. The trained policy is tested with and without masking the four constraint-related state features.}
		\label{fig:state_ablation}
	\end{figure}
	
	\noindent\textbf{State.} The state ablation masks the constraint-related entries $s_7$--$s_{10}$ during testing, while keeping the trained policy and SHADE optimizer unchanged. These entries describe violation progress, feasibility ratio, previous action, and pairwise objective-violation agreement. Fig.~\ref{fig:state_ablation} shows that masking them weakens 50D and 100D transfer, indicating that decision and objective statistics alone do not reveal the feasibility bottleneck reliably.
	
\begin{figure}[!t]
		\centering
		\includegraphics[width=0.78\linewidth]{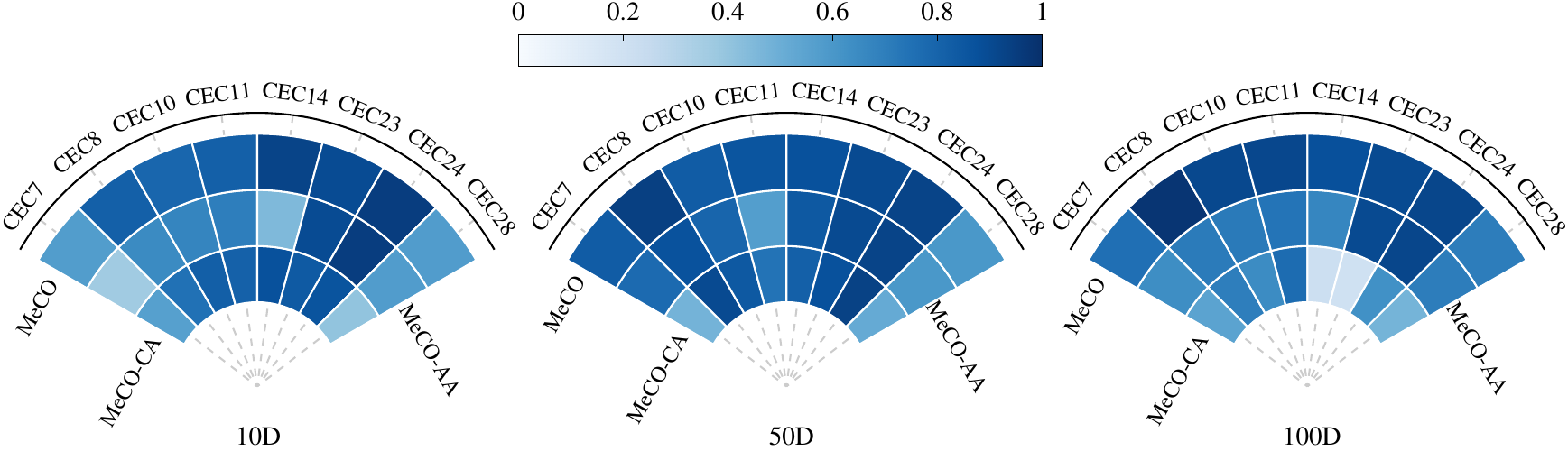}
		\caption{Action-space ablation on the fixed CEC2017 test set for 10D, 50D, and 100D. Darker colors denote better averaged final performance.}
		\label{fig:action_ablation}
	\end{figure}

	\noindent\textbf{Action.} The action ablation tests whether the geometric decoding rule is necessary for stable relaxation control. Using the fixed split, we train separate policies on 10D, 50D, and 100D for the original action design and two variants. Both variants replace Eq.~\eqref{eq:epsilon} with a linear multiplicative update
	\begin{equation}
		\boldsymbol{\epsilon}_t=\boldsymbol{\epsilon}_{t-1}(1-a_t).
		\label{eq:linear_ablation}
	\end{equation}
	The aggressive variant, MeCO-AA, can abruptly collapse or enlarge the relaxation vector; the conservative variant, MeCO-CA, often adapts too slowly. Fig.~\ref{fig:action_ablation} shows that the original geometric rule is more stable because Eq.~\eqref{eq:epsilon} interpolates between the initial violation scale and $\delta$, giving each action a consistent meaning across constraint magnitudes.
	
	\begin{figure}[!t]
		\centering
		\includegraphics[width=0.90\linewidth]{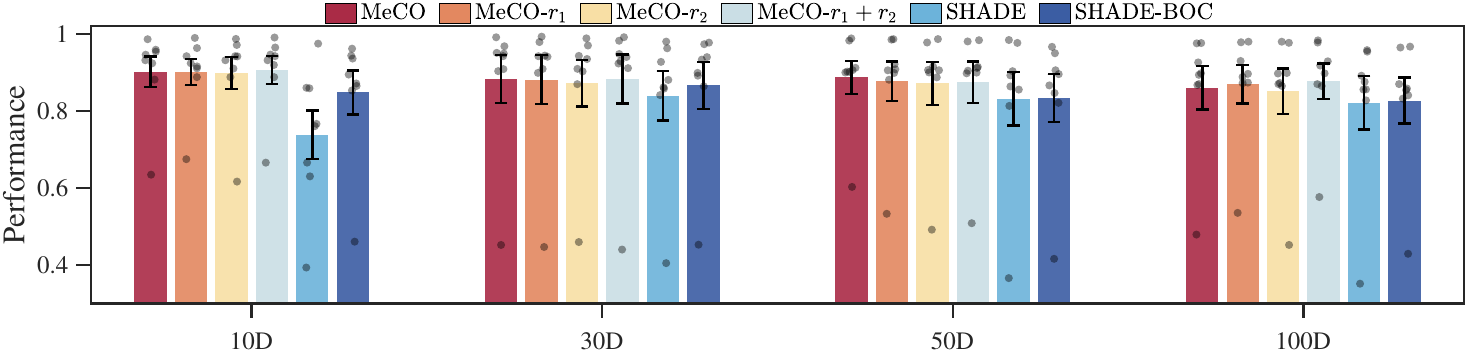}
		\caption{Reward ablation and baseline comparison on the fixed CEC2017 test set for 10D, 30D, 50D, and 100D. Higher values indicate better normalized relative performance.}
		\label{fig:reward_ablation}
	\end{figure}
	
	\noindent\textbf{Reward.} The reward ablation uses the same fixed split and compares full MeCO with three variants on 10D, 30D, 50D, and 100D: \emph{MeCO-$r_1$}, which uses only objective progress; \emph{MeCO-$r_2$}, which uses only violation reduction; and \emph{MeCO-$r_1+r_2$}, which directly sums both terms without the dynamic weight. Fig.~\ref{fig:reward_ablation} shows that the full reward is more stable because it avoids both infeasible low-objective bias and premature stagnation after feasibility is found.
	
	\begin{table}[!t]
		\centering
		\caption{Average relative reduction in $f(\mathbf{x}^\star)+\nu(\mathbf{x}^\star)$ for MeCO variants on the fixed CEC2017 test set. Each entry is averaged over $F_7$, $F_8$, $F_{10}$, $F_{11}$, $F_{14}$, $F_{23}$, $F_{24}$, and $F_{28}$, with the remaining functions used for training. Higher values are better.}
		\label{tab:training_gain}
		\scriptsize
		\setlength{\tabcolsep}{30pt}
		\renewcommand{\arraystretch}{1}
		\begin{tabular}{lcccc}
			\toprule
			Variant & 10D & 30D & 50D & 100D \\
			\midrule
			MeCO & \textbf{0.9406} & \textbf{0.9374} & \textbf{0.9251} & \textbf{0.9094} \\
			MeCO-u & 0.8877 & 0.8619 & 0.8439 & 0.8485 \\
			MeCO-rand & 0.8513 & 0.8927 & 0.8205 & 0.7736 \\
			\bottomrule
		\end{tabular}
	\end{table}
	
	\noindent\textbf{Meta-Training and Random Control.} The meta-training and random-control ablation tests whether the DDQN policy learns useful state-dependent behavior. Table~\ref{tab:training_gain} reports the average relative reduction in $f(\mathbf{x}^\star)+\nu(\mathbf{x}^\star)$ over the eight test functions, computed as
	\[
	1-\frac{f(\mathbf{x}_T^\star)+\nu(\mathbf{x}_T^\star)}
	{f(\mathbf{x}_0^\star)+\nu(\mathbf{x}_0^\star)}.
	\]
	Full MeCO obtains the highest gain in every dimension. MeCO-rand is occasionally competitive, but its lower gains at 10D, 50D, and 100D show that random actions are not a reliable substitute for learned state-dependent control.
	
	
	

	\subsection{Discussion}

	The experiments support two practical observations about learned relaxation control. First, MeCO is most useful when the search moves between infeasible exploration and feasible refinement: it can exploit nearly feasible candidates with good objective values and restore stricter feasibility pressure when violation becomes the bottleneck. Fixed penalties or fixed $\epsilon$ values cannot adapt this preference during optimization. The ES baselines also show that strong evolutionary search engines remain competitive, while MeCO offers a complementary way to learn transferable feasibility-pressure control.
	
	Second, transfer benefits from compact state and action abstractions. MeCO uses normalized population statistics and aggregate constraint-status features, including violation progress, feasibility ratio, previous action, and objective-violation agreement. These features provide constraint-sensitive information while keeping the controller fixed-dimensional across different problem dimensions and constraint numbers. Similarly, the scalar action is decoded according to constraint-specific violation scales, which gives the same action a consistent meaning across problems with different constraint magnitudes.
	
	These design choices trade fine-grained expressiveness for transferability. Future work can further enhance this framework with richer state encoders for detailed violation distributions and feasibility margins, vector-valued relaxation actions for per-constraint control, and stability-aware reward terms for more explicit action-smoothness modeling. Although the UAV tasks and engineering problems broaden the empirical scope, validation across more application domains is also an important direction.

	

	\section{Conclusion}
	This letter presents MeCO, a meta-learning-assisted framework for adaptive $\epsilon$-relaxation in constrained BBO. MeCO learns a DDQN policy that observes compact population and constraint features and controls the relaxation vector of a SHADE optimizer. Experiments on CEC2017, 16 UAV path-planning tasks, and eight real-world engineering design problems provide evidence that the learned policy transfers across held-out benchmark functions, higher dimensions, and application-domain tasks under low-budget online evaluation. The results also show that MeCO remains competitive against representative constrained BBO baselines, including ES-based methods. Future works should explore design refinement on  richer state encoders, vector-valued relaxation actions, stability-aware reward, as well as broader validation across additional constrained BBO applications.
	\appendix
	\setcounter{section}{0}
	\setcounter{equation}{0}
	\setcounter{figure}{0}
	\setcounter{table}{0}
	\renewcommand{\theequation}{\thesection.\arabic{equation}}
	\renewcommand{\thefigure}{\thesection.\arabic{figure}}
	\renewcommand{\thetable}{\thesection.\arabic{table}}
	\renewcommand{\theHequation}{\thesection.\arabic{equation}}
	\renewcommand{\theHfigure}{\thesection.\arabic{figure}}
	\renewcommand{\theHtable}{\thesection.\arabic{table}}
	\newcommand{\supplementaryincluded}{}
	\ifdefined\supplementaryincluded
	\providecommand{\method}{MeCO}
	\providecommand{\metric}{\ensuremath{f(\mathbf{x}^{\star})+\nu(\mathbf{x}^{\star})}}
	\newcommand{\suppsection}{\subsection}
	\newcommand{\suppsubsection}[1]{\noindent\textbf{#1.} }
	\section{Supplementary Material}
\else
	\pdfminorversion=7
	\documentclass[letterpaper,journal]{IEEEtran}
	\usepackage{amsmath,amsfonts}
	\usepackage{array}
	\usepackage{booktabs}
	\usepackage{cite}
	\usepackage{graphicx}
	\usepackage{makecell}
	\usepackage{multirow}
	\usepackage{stfloats}
	\usepackage{url}
	\usepackage[table]{xcolor}

	\newcommand{\method}{MeCO}
	\newcommand{\metric}{\ensuremath{f(\mathbf{x}^{\star})+\nu(\mathbf{x}^{\star})}}
	\newcommand{\suppsection}{\section}
	\newcommand{\suppsubsection}[1]{\noindent\textbf{#1.} }

	\begin{document}
	\title{Supplementary Material for ``Meta-Learning-Assisted Constraint Relaxation for Constrained Black-Box Optimization''}
	\maketitle
\fi
	
	\suppsection{UAV Transfer Experiment}
	\label{sec:supp_uav}
	
	
	
	\suppsubsection{UAV Problem Description}
	\label{sec:supp_uav_problem}
	The UAV task is based on wireless-sensor data collection by a rotary-wing UAV~\cite{bai2025uav}, and is extended here with explicit geometric obstacles to form a constrained black-box testbed. The UAV flies at a fixed altitude $H$ over a two-dimensional area $\mathcal{A}=[0,U_x]\times[0,U_y]$ and collects data from $J$ ground sensors $\mathbf{S}=\{\mathbf{s}_1,\ldots,\mathbf{s}_J\}$. A candidate trajectory is represented by the ordered hovering points $\mathbf{P}=\{\mathbf{p}_1,\ldots,\mathbf{p}_I\}$ and the corresponding hovering durations $\mathbf{T}_{hov}=\{T_{hov,1},\ldots,T_{hov,I}\}$, where $I=6$ for all instances. We set $\mathbf{p}_0=\mathbf{p}_{start}$ and $\mathbf{p}_{I+1}=\mathbf{p}_{end}$. For instances~\#01--\#12, the start and end points are distinct fixed points; for instances~\#13--\#16, the UAV is required to return to its starting point, i.e., $\mathbf{p}_{end}=\mathbf{p}_{start}$.
	
	For each consecutive pair $\mathbf{p}_i$ and $\mathbf{p}_{i+1}$, $i=0,\ldots,I$, let $d_i=\|\mathbf{p}_{i+1}-\mathbf{p}_i\|_2$. The flight time follows a trapezoidal velocity model:
	\begin{equation}
		T_{fly,i} =
		\begin{cases}
			\frac{d_i}{v_{max}}+\frac{v_{max}}{a}, & d_i>\frac{v_{max}^2}{a},\\
			2\sqrt{\frac{d_i}{a}}, & d_i\leq\frac{v_{max}^2}{a}.
		\end{cases}
		\label{eq:supp_flight_time}
	\end{equation}
	When hovering at $\mathbf{p}_i$, the transmission rate from sensor $j$ is
	\begin{equation}
		R_{i,j}=\max\{R_c-\xi\sqrt{\|\mathbf{p}_i-\mathbf{s}_j\|_2^2+H^2},0\},
		\label{eq:supp_rate}
	\end{equation}
	and the total data collected from sensor $j$ is $Q_j=\sum_{i=1}^{I}R_{i,j}T_{hov,i}$. To model safety constraints, the map contains rectangular obstacles, circular obstacles, and polygonal no-fly zones. Let $\mathcal{O}$ denote their union and let $\mathcal{S}_i$ be the line segment from $\mathbf{p}_i$ to $\mathbf{p}_{i+1}$ for $i=0,\ldots,I$. The collision measure is
	\begin{equation}
		L_{vio}(\mathbf{P})=\sum_{i=0}^{I}\mathrm{len}(\mathcal{S}_i\cap\mathcal{O}),
		\label{eq:supp_collision_length}
	\end{equation}
	where $\mathrm{len}(\cdot)$ denotes Euclidean length. Thus, $L_{vio}(\mathbf{P})$ is zero only when the whole path avoids all restricted regions, and otherwise measures the total distance traveled inside obstacles or no-fly zones. The resulting constrained optimization problem is
	\begin{equation}
		\label{eq:supp_uav_problem}
		\begin{aligned}
			\min_{\mathbf{x}} \quad
			& \alpha\sum_{i=0}^{I}T_{fly,i}+(1-\alpha)\sum_{i=1}^{I}T_{hov,i},\\
			\mathrm{s.t.}\quad
			& \mathbf{p}_i\in\mathcal{A},\quad 0\leq T_{hov,i}\leq T_{max},\\
			& \sum_{j=1}^{J}R_{i,j}\leq R_{\lambda},\quad Q_j\geq Q_{\lambda},\\
			& L_{vio}(\mathbf{P})=0.
		\end{aligned}
	\end{equation}
	Here $\mathbf{x}=\{\mathbf{P},\mathbf{T}_{hov}\}$. The constraints enforce spatial bounds, hovering-time limits, communication bandwidth, sufficient data collection, and collision-free flight.
	
	In the experiments, infeasible paths are penalized according to the violated constraints. In particular, $L_{vio}(\mathbf{P})$ directly quantifies obstacle violation by the total path length inside restricted regions, so a smaller value indicates less severe collision risk and $L_{vio}(\mathbf{P})=0$ indicates a collision-free trajectory.
	
	Table~\ref{tab:supp_uav_parameters} lists the parameters used for all UAV instances. These settings are the same as those used in the main paper.
	
	\begin{table}[!t]
		\centering
		\caption{Parameters of the UAV mission environment.}
		\label{tab:supp_uav_parameters}
		\scriptsize
		\setlength{\tabcolsep}{3pt}
		\renewcommand{\arraystretch}{1.12}
		\begin{tabular}{p{1.05cm}p{1.35cm}p{4.65cm}}
			\toprule
			Parameter & Value & Description \\
			\midrule
			$I$ & $6$ & Number of hovering points. \\
			$v_{max}$ & $40\,\mathrm{m/s}$ & Maximum operational UAV speed. \\
			$a$ & $1\,\mathrm{m/s^2}$ & UAV acceleration. \\
			$T_{max}$ & $200\,\mathrm{s}$ & Maximum hovering time at each point. \\
			$H$ & $20\,\mathrm{m}$ & UAV flying altitude. \\
			$\xi$ & $0.025$ & Signal attenuation factor. \\
			$R_c$ & $50\,\mathrm{Mbit/s}$ & Central data rate of a sensor. \\
			$R_{\lambda}$ & $200\,\mathrm{Mbit/s}$ & Maximum aggregate receiving data rate of the UAV. \\
			$Q_{\lambda}$ & $500\,\mathrm{Mbit}$ & Required data volume from each sensor. \\
			\bottomrule
		\end{tabular}
	\end{table}
	
	\begin{figure}[!t]
		\centering
		\includegraphics[width=0.85\linewidth]{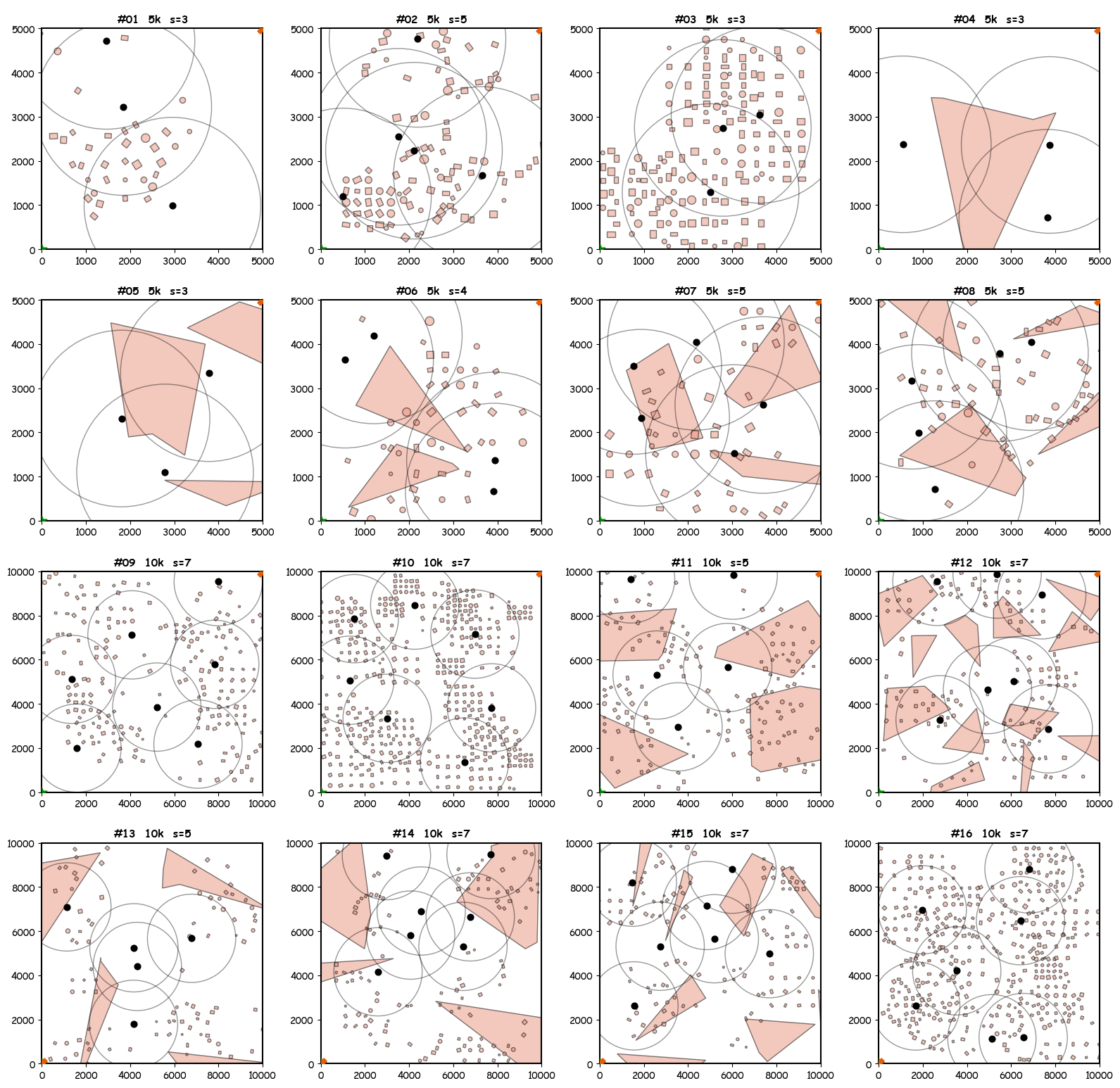}
		\caption{The 16 UAV path-planning instances used in the application-domain transfer experiment. The maps include different area sizes, sensor distributions, hovering requirements, rectangular and circular obstacles, and polygonal no-fly zones. For instances~\#13--\#16, the UAV must return to its starting point after visiting the hovering points.}
		\label{fig:supp_uav_instances}
	\end{figure}
	
	\suppsubsection{Normalized UAV Performance}
	\label{sec:supp_uav_performance}
	The UAV boxplots use a normalized performance score to make the 16 UAV instances comparable on a common scale. For algorithm $a$ on UAV instance $p$ and independent run $r$, the raw score is
	\begin{equation}
		s_{p,a,r}=f(\mathbf{x}^{\star}_{p,a,r})+\nu(\mathbf{x}^{\star}_{p,a,r}),
		\label{eq:supp_uav_raw_score}
	\end{equation}
	where lower values are better. For each UAV instance $p$, we first identify the best and worst observed raw scores over all compared algorithms and all independent runs:
	\begin{equation}
		s^{\mathrm{best}}_{p}=\min_{a,r}s_{p,a,r}, \qquad
		s^{\mathrm{worst}}_{p}=\max_{a,r}s_{p,a,r}.
		\label{eq:supp_uav_best_worst}
	\end{equation}
	The plotted performance is then
	\begin{equation}
		\mathrm{Performance}_{p,a,r}=
		\begin{cases}
			\dfrac{s^{\mathrm{worst}}_{p}-s_{p,a,r}}
			{s^{\mathrm{worst}}_{p}-s^{\mathrm{best}}_{p}}, & s^{\mathrm{worst}}_{p}>s^{\mathrm{best}}_{p},\\[1.2ex]
			1, & s^{\mathrm{worst}}_{p}=s^{\mathrm{best}}_{p}.
		\end{cases}
		\label{eq:supp_uav_perf}
	\end{equation}
	This form makes the direction explicit: the best observed raw score on the same UAV instance receives $\mathrm{Performance}=1$, and the worst observed raw score receives $\mathrm{Performance}=0$. This normalization is used only for cross-instance visualization; the underlying optimization objective and final evaluation remain based on the raw \metric.
	
\begin{figure}[!t]
		\centering
		\includegraphics[width=0.90\linewidth]{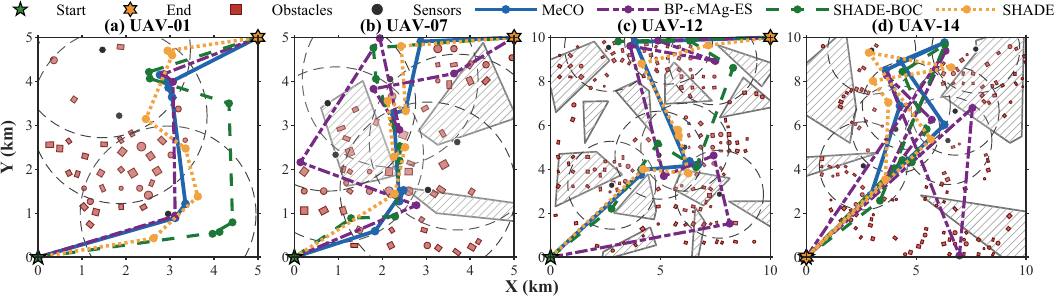}
		\caption{Representative UAV trajectories obtained by selected compared methods on four selected instances. For each method, the trajectory is selected by lexicographically minimizing $(v(\mathbf{x}^{\star}), f(\mathbf{x}^{\star}))$.}
		\label{fig:supp_uav_trajectories}
	\end{figure}

	\suppsubsection{UAV Results}
	\label{sec:supp_uav_results}
	Figure~\ref{fig:supp_uav_instances} shows the 16 tested UAV instances, and Fig.~\ref{fig:supp_uav_trajectories} visualizes representative trajectories.
	The UAV results show that \method~maintains competitive transfer performance across the 16 path-planning instances.
	Since the policy is trained only on the CEC2017 constrained benchmark and is directly applied to UAV path planning without UAV-specific training, these results provide additional evidence that the learned relaxation-control strategy can generalize beyond synthetic benchmark functions.
	
	Figure~\ref{fig:supp_uav_trajectories} presents representative UAV trajectories on UAV-01, UAV-07, UAV-12, and UAV-14.
	Despite the differences in obstacle layouts and sensor distributions, \method~is able to generate feasible and competitive trajectories across the selected instances.
	For UAV-01, UAV-07, and UAV-12, \method~successfully navigates the obstacle fields while maintaining relatively compact routes.
	For UAV-14, which involves a return-flight task with a more complex trajectory structure, \method~also finds a feasible route through the densely constrained environment.
	Overall, these trajectory examples complement the quantitative results and further demonstrate the effectiveness of transferring MeCO from synthetic CEC2017 problems to practical UAV path-planning tasks.

	\begin{table}[!t]
		\centering
		\caption{Performance on eight additional real-world constrained optimization problems. Each cell reports the mean objective (top) and standard deviation (bottom) over feasible runs; infeasible methods report ``--'' and the mean violation. Rankings follow the feasibility-first rule. Gray cells indicate the best and second-best results.}
		\label{tab:supp_real_world}
		\tiny
		\setlength{\tabcolsep}{2.2pt}
		\renewcommand{\arraystretch}{1.10}
		\resizebox{\linewidth}{!}{%
			\begin{tabular}{lcccccccc}
				\toprule
				Problem & \method & SHADE & SHADE-$\epsilon$ & SHADE-BOC & SACOBRA & NSGA-III & BP-$\epsilon$MAg-ES & CMA-ES \\
				\midrule
				RC19 & \cellcolor{gray!8}\makecell{1.72189e+00\\9.96224e-03} & \makecell{1.73993e+00\\2.69459e-02} & \makecell{3.22954e+00\\8.22563e-01} & \makecell{1.80046e+00\\7.78641e-02} & \makecell{1.91641e+00\\7.37513e-02} & \makecell{2.79471e+00\\9.29770e-01} & \makecell{2.12196e+00\\3.90596e-01} & \cellcolor{gray!18}\textbf{\makecell{1.67913e+00\\8.41412e-03}} \\
				RC20 & \cellcolor{gray!8}\makecell{2.63902e+02\\1.25689e-02} & \makecell{2.73264e+02\\4.49794e+00} & \makecell{2.71863e+02\\2.96576e+00} & \makecell{2.63906e+02\\1.45156e-02} & \makecell{2.65495e+02\\9.27442e-01} & \makecell{2.66598e+02\\2.33792e+00} & \cellcolor{gray!18}\textbf{\makecell{2.63897e+02\\7.64763e-03}} & \makecell{2.74539e+02\\9.71506e+00} \\
				RC23 & \cellcolor{gray!8}\makecell{--\\1.54290e-02} & \makecell{--\\3.16377e-02} & \makecell{--\\3.71041e+01} & \makecell{--\\4.47136e-02} & \makecell{--\\3.52672e+01} & \makecell{--\\3.40029e+00} & \cellcolor{gray!18}\textbf{\makecell{1.66050e+01\\3.18481e-01}} & \makecell{--\\3.47942e-02} \\
				RC24 & \makecell{5.72190e+00\\2.46685e-01} & \makecell{6.54315e+00\\6.83324e-01} & \makecell{9.88278e+01\\1.85544e+02} & \makecell{6.36841e+00\\6.04817e-01} & \cellcolor{gray!8}\makecell{5.66621e+00\\1.61578e+00} & \makecell{5.69190e+00\\1.19672e+00} & \makecell{1.40024e+03\\4.32631e+03} & \cellcolor{gray!18}\textbf{\makecell{4.14008e+00\\2.78553e-01}} \\
				RC25 & \cellcolor{gray!18}\textbf{\makecell{2.64888e+03\\2.47055e+02}} & \cellcolor{gray!8}\makecell{2.98596e+03\\5.51026e+02} & \makecell{4.70318e+03\\4.83030e+02} & \makecell{3.72447e+03\\9.04759e+02} & \makecell{4.25680e+03\\3.60229e+02} & \makecell{4.98999e+03\\1.03032e+03} & \makecell{5.48099e+03\\2.42768e+03} & \makecell{3.18939e+03\\1.27894e+03} \\
				RC27 & \cellcolor{gray!18}\textbf{\makecell{5.45637e+02\\4.35766e+00}} & \makecell{7.31415e+02\\6.75375e+01} & \makecell{6.53984e+02\\6.40050e+01} & \cellcolor{gray!8}\makecell{5.47662e+02\\6.54271e+00} & \makecell{5.48374e+02\\7.48924e+00} & \makecell{5.93166e+02\\2.32898e+01} & \makecell{5.72933e+02\\2.06837e+01} & \makecell{--\\1.02112e-01} \\
				RC29 & \cellcolor{gray!18}\textbf{\makecell{2.98164e+06\\6.03042e+03}} & \makecell{5.83551e+06\\1.12088e+06} & \makecell{3.76269e+06\\5.73176e+05} & \cellcolor{gray!8}\makecell{3.03841e+06\\4.13052e+04} & \makecell{3.19362e+06\\9.77764e+04} & \makecell{3.06387e+06\\9.20691e+04} & \makecell{3.16667e+06\\2.08226e+05} & \makecell{4.42519e+06\\8.96857e+05} \\
				RC32 & \makecell{-3.05871e+04\\1.20646e+01} & \makecell{-2.95183e+04\\3.51585e+02} & \makecell{-3.03478e+04\\1.72476e+02} & \makecell{-3.05057e+04\\3.69550e+01} & \cellcolor{gray!18}\textbf{\makecell{-3.06276e+04\\2.16692e+01}} & \makecell{-3.04427e+04\\1.02901e+02} & \cellcolor{gray!8}\makecell{-3.06172e+04\\9.20678e+01} & \makecell{-2.96622e+04\\2.82853e+02} \\
				\midrule
				Avg. rank & \cellcolor{gray!18}\textbf{\makecell{2.000\\(1)}} & \makecell{5.500\\(7)} & \makecell{6.625\\(8)} & \cellcolor{gray!8}\makecell{3.625\\(2)} & \makecell{4.000\\(3)} & \makecell{5.125\\(6)} & \makecell{4.250\\(4)} & \makecell{4.875\\(5)} \\
				\bottomrule
			\end{tabular}%
		}
	\end{table}
    
	\suppsection{Additional Real-World Problems}
	\label{sec:supp_real_world}
	
	To broaden the real-world evaluation beyond the UAV application, we further evaluate MeCO and the comparison methods on eight constrained engineering design problems selected from the real-world test suite~\cite{kumar2020test}. These problems span a diverse range of engineering scenarios, from structural and mechanical design to process-system optimization, and involve different dimensionalities and constraint structures, including both inequality and equality constraints. As in the UAV experiments, MeCO is directly applied to these problems without problem-specific retraining or fine-tuning. The same objective convention, final violation measure, and evaluation protocol as in the main paper are adopted, with SACOBRA and NSGA-III additionally included alongside the ES baselines. Table~\ref{tab:supp_real_world_problem_summary} summarizes the selected problems, and Table~\ref{tab:supp_real_world} reports the corresponding optimization results.
	
	\begin{table}[!t]
		\centering
		\caption{Summary of the eight additional real-world constrained optimization problems. $D$ is the number of decision variables, $g$ is the number of inequality constraints, $h$ is the number of equality constraints, and $f_{\mathrm{best}}^{\mathrm{known}}$ denotes the best-known objective value reported for the problem.}
		\label{tab:supp_real_world_problem_summary}
		\tiny
		\setlength{\tabcolsep}{2.2pt}
		\renewcommand{\arraystretch}{1.10}
		\resizebox{0.8\columnwidth}{!}{
			\begin{tabular}{clcccc}
				\toprule
				ID & Problem name & $D$ & $g$ & $h$ & $f_{\mathrm{best}}^{\mathrm{known}}$ \\
				\midrule
				RC19 & Welded beam design~\cite{he2018covariance_de} & 4 & 5 & 0 & $1.67022\mathrm{E}{+00}$ \\
				RC20 & Three-bar truss design~\cite{shadravan2019sailfish} & 2 & 3 & 0 & $2.63896\mathrm{E}{+02}$ \\
				RC23 & Step-cone pulley design~\cite{rao2009engineering} & 5 & 8 & 3 & $1.60699\mathrm{E}{+01}$ \\
				RC24 & Robot gripper design~\cite{osyczka1999robot_gripper} & 7 & 7 & 0 & $2.52879\mathrm{E}{+00}$ \\
				RC25 & Hydro-static thrust bearing design~\cite{siddall1982optimal} & 4 & 7 & 0 & $1.62544\mathrm{E}{+03}$ \\
				RC27 & 10-bar truss design~\cite{hohuu2016truss} & 10 & 3 & 0 & $5.24451\mathrm{E}{+02}$ \\
				RC29 & Gas transmission compressor design~\cite{beightler1976geometric} & 4 & 1 & 0 & $2.96490\mathrm{E}{+06}$ \\
				RC32 & Himmelblau's function~\cite{himmelblau1972nonlinear} & 5 & 6 & 0 & $-3.06655\mathrm{E}{+04}$ \\
				\bottomrule
			\end{tabular}
		}
	\end{table}
		
	The detailed optimization results and overall average ranks are reported in Table~\ref{tab:supp_real_world}. Overall, \method~achieves the best average rank of 2.000 across the eight additional engineering problems. It obtains the best results on RC25, RC27, and RC29, and ranks among the top two methods on five of the eight problems, indicating consistently competitive performance across different engineering scenarios. Among the ES baselines, BP-$\epsilon$MAg-ES is the only method that finds feasible solutions on the challenging RC23 problem, while CMA-ES achieves the best results on RC19 and RC24. Together with the UAV experiments, these results extend the evaluation beyond a single application domain and show that \method~maintains strong overall performance across diverse constrained engineering problems.
	
\ifdefined\supplementaryincluded
\else
	\bibliographystyle{IEEEtran}
	\bibliography{refers}
\end{document}
\fi

	\bibliographystyle{IEEETran}
	\bibliography{refers}
	
\end{document}